\documentclass[lettersize,journal]{IEEEtran}
\usepackage{amsmath,amsfonts}
\usepackage{algorithmic}
\usepackage{array}
\usepackage{textcomp}
\usepackage{stfloats}
\usepackage{url}
\usepackage{verbatim}
\usepackage{threeparttable}
\usepackage{graphicx}
\usepackage{svg}
\usepackage{cite}
\usepackage[hidelinks]{hyperref}
\def\BibTeX{{\rm B\kern-.05em{\sc i\kern-.025em b}\kern-.08em
		T\kern-.1667em\lower.7ex\hbox{E}\kern-.125emX}}
\usepackage{balance}
\usepackage{booktabs}
\usepackage{xcolor}
\usepackage{ulem} 
\usepackage[thicklines]{cancel} 
\ifCLASSOPTIONcompsoc
  \usepackage[caption=false,font=normalsize,labelfont=sf,textfont=sf]{subfig}
\else
  \usepackage[caption=false,font=footnotesize]{subfig}

\usepackage{bm}

\begin{document}
	\title{\textcolor{black}{Coordination for Connected and Automated Vehicles at Non-signalized Intersections: A Value Decomposition-based Multiagent Deep Reinforcement Learning Approach}}
	\author{Zihan Guo, Yan Wu, Lifang Wang, Junzhi Zhang
        \thanks{Copyright (c) 2015 IEEE. Personal use of this material is permitted. However, permission to use this material for any other purposes must be obtained from the IEEE by sending a request to pubs-permissions@ieee.org.}
        \thanks{This work is financially supported by the Institute of Electrical Engineering Chinese Academy of Sciences (E1553301).}
		\thanks{Z. Guo, Y. Wu, L. Wang are with the Key Laboratory of Power Electronics and Electric Drives, Institute of Electrical Engineering Chinese Academy of Sciences, Beijing 100190, China and also with University of Chinese Academy of Sciences, Beijing 100190, China. J. Zhang is with the State Key Laboratory of Automotive Safety and Energy, Department of Automotive Engineering, Tsinghua University, Beijing 100084, China. ({\tt\small Email: zh\_guo97@163.com; wuyan@mail.iee.ac.cn; wlf@mail.iee.ac.cn; jzhzhang@mail.tsinghua.edu.cn})}
	    \thanks{*The corresponding author is Yan Wu and Lifang Wang. All questions about this paper should be sent to email {\tt\small wuyan@mail.iee.ac.cn} and {\tt\small wlf@mail.iee.ac.cn}}
     \thanks{Digital Object Identifier 10.1109/TVT.2022.3219428}
     }

	\maketitle

	\begin{abstract}
		The recent proliferation of the research on multi-agent deep reinforcement learning (MDRL) offers an encouraging way to coordinate multiple connected and automated vehicles (CAVs) to pass the intersection. In this paper, we apply a value decomposition-based MDRL approach (QMIX) to control various CAVs in mixed-autonomy traffic of different densities to efficiently and safely pass the non-signalized intersection with fairish fuel consumption. Implementation tricks including network-level improvements, Q value update by TD ($\lambda$), and reward clipping operation are added to the pure QMIX framework, which is expected to improve the convergence speed and the asymptotic performance of the original version. The efficacy of our approach is demonstrated by several evaluation metrics: average speed, the number of collisions, and average fuel consumption per episode. The experimental results show that our approach's convergence speed and asymptotic performance can exceed that of the original QMIX and the proximal policy optimization (PPO), a state-of-the-art reinforcement learning baseline applied to the non-signalized intersection. Moreover, CAVs under the lower traffic flow controlled by our method can improve their average speed without collisions and consume the least fuel. The training is additionally conducted under the doubled traffic density, where the learning reward converges. Consequently, the model with maximal reward and minimum crashes can still guarantee low fuel consumption, but slightly reduce the efficiency of vehicles and induce more collisions than the lower-traffic counterpart, implying the difficulty of generalizing RL policy to more advanced scenarios.
	\label{abs}
	\end{abstract}
	

	\section{Introduction}
	\label{intro}
        \IEEEPARstart{T}{he} traffic congestion and potential accidents in urban areas result from the growing urbanized population and the resulting rise in vehicle demands. It has imposed challenges to the modern transportation system.  In 2014, people in cities suffered 6.9 billion hours more wasted time on the road, and the extra fuel consumption was about 3 billion gallons.\cite{RN11}. Fortunately, CAVs-related technologies, such as vehicle-to-vehicle communication (V2V) and vehicle-to-infrastructure communication(V2I), empower vehicles to expand their line of sight to obtain more information about the surroundings, e.g., the position, speed of all cars and other road participants within the communication range of road infrastructures \cite{RN16}. Consequently, it is possible and beneficial for vehicles to cooperate with others to acquire safe and efficient traffic. 
    
	    According to the literature, the most straightforward attempt to manage the traffic at an intersection is to build traffic lights. For instance, the manual \cite{RN147} uses a pre-defined signal phase and timing (SPaT) plan to handle steady traffic conditions. P. Varaiya \cite{varaiya2013max} designs a greedy policy such that the infrastructure can select the signal phase maximizing the pressure. However, these schemes still have several limitations: 1) they cannot eliminate the stop delay of vehicles during the red phase; 2) if the traffic of each incoming approach distributes equally, the overall efficacy could be undermined \cite{RN33}. Therefore, one branch develops adaptive traffic signal control (ATSC) strategies based on deep reinforcement learning (DRL). H. Wei, et al., \cite{RN32} propose a mix of online and offline RL framework based on deep Q network (DQN) with the state defined by queue length, waiting time, traffic signal phase, etc., discrete action determining if the traffic light can change its phase, a reward function including the sum of queue length and delay, etc.	
	
	    Another line of work is to forsake traffic signals and employ the RL framework to control one or more vehicles. The authors of the paper\cite{RN42} train an RL agent based on deep Q-learning. The goal is to navigate an ego vehicle with limited sensing ability to pass the road. The action space is represented in three different macro ways (time-to-go, sequential actions, and creep-and-go), and the empirical results show that the DQN agent can beat the rule-based time-to-collision (TTC) method in most scenarios. Nonetheless, one of the drawbacks is that a specific car-following model describes the behavior of the surrounding cars not controlled by DQN, and hence behaviors out of the scope of such a model are not well-comprehended by RL-controlled vehicles. In research \cite{RN33}, a gradient-based RL agent (i.e., proximal policy optimization, with abbreviation PPO) is trained with the help of a fixed environment model. The simulation results are more sample efficient and robust to the environmental noise as appropriate imagination depth is selected than the pure PPO approach and more computationally efficient than the conventional centralized optimization-based algorithm. Moreover, E. Vinitsky et al.\cite{vinitsky2018benchmarks} propose a FLOW framework, a benchmark for applying RL in mixed-autonomy scenarios, including intersections with and without traffic signals, merging, and bottleneck scenarios. Furthermore, they also test the performances of two RL algorithms, Trust Region Policy Optimization (TRPO) and PPO, and the other two non-RL baselines. These RL approaches are better solutions for merging scenarios. B. Peng et al.\cite{2021Connected} expand the figure-eight scenario in the FLOW benchmark to a two-lane non-signalized intersection and control two CAVs to reduce the queuing phenomenon. Besides, the enthusiasm of researchers applying DRL or deep learning (DL) methods to the scenarios beyond the intersection, such as ramp metering and estimating the traffic incident duration, is growing \cite{zhu2021dynamic, belletti2017expert}.

	    However, directly deploying single-agent RL framework to multi-agent cooperative setting can cause the non-stationarity problem \cite{RN47}. Intended to conquer this problem, a centralized training and decentralized execution (CTDE) framework appears in the field of multi-agent deep reinforcement learning (MDRL). Firstly, the decentralization scheme is easy to implement but brings about instability to learning as the environment is dynamically changing with other agents' learning and explorations. On the other hand, it is granted that the centralization architecture can overcome the non-stationarity, splicing all agents’ joint actions and observations impairs its scalability to large-scale applications when thousands of agents are involved \cite{RN48}. The basic procedure of the CTDE framework is to use the global states for training and to deploy the learned model in a decentralized fashion when the training has been completed. One instantiation of the CTDE framework is to employ the value decomposition mechanism. One mode is to decompose the total fitted Q value directly into the sum of each agent's approximate Q value during training, called the value decomposition networks (VDNs) \cite{sunehag2018vdn}. However, VDN does not fully exploit the benefits of adding global states during training, which has potential to increase the representing power of joint action-value functions. Therefore, QMIX \cite{RN189} creatively employs a monotonic constraint to achieve better expressivity of the centralized action-value function. Although a band of research endeavors to relax the monotonicity constraint of QMIX theoretically \cite{Qtran,yang2020multi}, a recent study \cite{RN54} manifests that QMIX can gain the state-of-the-art average score at StarCraft Multi-agent Challenge (SMAC) task \cite{samvelyan2019starcraft} after fine-tuning with some code-level tricks. Another important work is based on policy gradient to solve a multi-agent continuous control task (MADDPG) \cite{RN52}. However, the computational complexity of MADDPG architecture increases with the number of agents, which is detrimental to its scalability.
	    
	    The key contributions of our work are: 1) Adding several implementation tricks to the original QMIX algorithm, including employing the eligibility traces, reward clipping, and network-level improvements. 2) Using modified QMIX to control multiple vehicles (8 vehicles in total) at the non-signalized intersection with four road segments and eight lanes in mixed-autonomy traffic, where human-driven vehicles (HDVs) are simulated by Intelligent Driver Model(IDM). 3) Compared to the PPO algorithm, our approach can improve the average speed, average delayed time, and average fuel consumption and simultaneously assure safety.
	    
	    The rest of this paper is arranged as follows. Section \ref{prel}. gives a concise introduction of the necessary background knowledge of DRL and principles behind the QMIX approach, and Section \ref{formulation}. transforms the coordination problem of multiple CAVs at the intersection into an RL framework. Section \ref{methods}. elaborates on the control flow of the QMIX algorithm and code-level tricks we employed for fine-tuning. Section \ref{results}. illustrates the simulation settings and exhibits the empirical results. Section \ref{conclusion}. provides the summary and the works in the future.

	\section{Preliminaries}
	\label{prel}
	
	\subsection{Decentralized POMDP}
	\noindent
	The decentralized partially observable Markov decesion process(Dec-POMDP) \cite{RN149} describes a fully cooperative multiagent setting with partial observation. It is defined as a tuple $G = \left( {I,S,\{ {A_i}\} ,T,r,\{ {\Omega _i}\} ,O,\gamma } \right)$, in which $I\in\mathbb{R}^N$ is the finite set of all agents contained in the problem, $s\in S$ denotes the true state of the environment. $A_i$ is a finite set of actions of agent $i\in I$, and the space of joint actions is defined as $A=\times_i A_i$ ($\times$ is the operator of Cartesian product) and $\hat{a}\in A$ represents the joint actions taken by all agents. $T:S \times A \times S \to \mathbb{R}$ is a state transition probability function, which is defined as $T(s,\bm{a},s') = \Pr (s'|\bm{a},s)$. A reward function $r:S\times A\times S\to \mathbb{R}$ is a scalar received by the team of all agents, which means the individual reward cannot be offered in Dec-POMDP setting. $\Omega_i$ is denoted as a finite set of observations for each agent $i\in I$ and correspondingly $\Omega=\times_i\Omega_i$ the set of joint observations. An observation probability function is $O:S \times A \times \Omega  \to [0,1]$ with the joint actions and observations (where $o_i\in \Omega_i$ and $\bm{o}\in \Omega$) as independent variables, which is subtly different from that in POMDP. Hence, $O(\bm o,\bm a,s') = {\rm{Pr}}(\bm o|\bm a,s')$. The last element $\gamma$ is described as the discount factor.
	
	\subsection{Deep Q-learning}
	\noindent
	Deep Q-learning algorithm \cite{mnih2015human} utilizes an approximate function parameterized by $\theta$ to represent the action-value function $Q(\cdot)$ and employs a replay buffer to store the transition data $(s,a,r,s')$ at each time step for the purpose of stabilizing the policy training, where the action $a$ taken by the agent in state $s$ leads to next state $s'$. To update the parameter of deep Q networks (DQNs), a mini-batch with size $b$ is sampled from the replay buffer, and the following loss function based on TD-error is minimized:
	\begin{equation}
	{\cal L}(\theta ) = \mathop \sum \limits_{i = 1}^b \left[ {{{\left( {y_i^{{\rm{target}}} - Q(s,a;\theta )} \right)}^2}} \right] \label{eq3}
	\end{equation}
	where ${y^{{\rm{target}}}} = {r_t} + \gamma {\max _{a'}}Q(s',a';\theta ')$. $\theta'$ is the parameter of the target network that replicates the network $\theta$ for a fixed period and keeps unvarying.
 
	\subsection{Independent Q-learning and VDNs}
		
    \noindent
    One straightforward way to solve a multi-agent problem is to transform it into a combination of single-agent problems in which each agent resides in the same environment. One of the earliest methods with this idea is called independent Q-learning (IQL) \cite{tan1993multi}. However, the concurrently and constantly changing policies of multiple agents force the IQL to experience the non-stationarity problem, and consequently, the convergence of this algorithm cannot be guaranteed. Nonetheless, the performance of IQL is not overshadowed by this risk of divergence, and when researchers tackle with cooperative or competitive games, IQL becomes a common benchmark \cite{tampuu2017multiagent}.
	
	Conversely, Sunehag et al. comes up with an MDRL algorithm called VDNs intended to utilize individual action-value function $Q_i(\tau_i,a_i;\theta_i)$ with local trajectory $\tau_i$, local action $a_i$, and parameters $\theta_i$ to estimate a joint value function $Q_{tot}(\bm{\tau}, \bold a)$, where $\bm{\tau} \in \bold{\cal T}^N$ is a joint action-observation history trajectory, and $\bm{a}\in A^N$ is a joint action. The decomposition formula is shown as follows.
	\begin{equation}
	    Q_{tot}(\bm{\tau},\bm{a})=\sum_{i=1}^N Q_i(\tau_i,a_i;\theta_i)
	\label{VDN_sum}
	\end{equation}
	The loss function of VDNs is the same as eq. (\ref{eq3}) except that $Q$ is substituted by $Q_{tot}$. The convenience of the value decomposition mechanism is that it can avoid the combinatorial nature of multi-agent problems since each value function approximator only employs the local trajectory and action of each learning agent \cite{RN48}. 
	
	\subsection{The Key Insights of QMIX}
	\noindent
	There are two key aspects of the QMIX framework, consisting of the individual-global-max (IGM) principle \cite{Qtran} and the monotonicity constraint proposed in QMIX.
	
	The IGM principle states as follows. The optimal joint action derived from the global maximization of the joint action-value function ${Q_{{\rm{jt}}}}:{{\cal T}^N} \times {A^N} \to \mathbb{R}$ (where $\bm{\tau} \in \cal T^N$ is a joint action-observation history trajectory) of a team of agents is equivalent to a collection of individually maximized action-value function $[{Q_i}:{\cal T} \times A \to \mathbb{R}]_{i = 1}^N$.
    
    \begin{equation}
	\mathop {\arg \max }\limits_{\bf{a}} {Q_{tot}}({\bm{\tau }},{\bm{a}}) = \left( \begin{array}{l}
	\arg {\max}_{{a_1}} {Q_1}({\tau _1},{a_1})\\
	\quad \quad \quad \;\; \vdots \quad \quad \\
	\arg {\max _{{a_N}}}{Q_N}({\tau _N},{a_N})
	\end{array} \right) \label{eq4}
	\end{equation}

	Moreover, the monotonicity constraint is one sufficient condition of eq. (\ref{eq4})
	\begin{equation}
	\frac{{\partial {Q_{tot}}({\bm{\tau }},{\bf{a}})}}{{\partial {Q_i}({\tau _i},{a_i})}} \ge 0,\quad \forall i \in I \label{eq5}
	\end{equation}
	
	A thorough procedure of QMIX and some implementation tricks added to it in our work is illustrated in Section \ref{methods}.  
	
	\section{Problem Statement and Formulation}
	\label{formulation}
	\subsection{Problem Statement}
\begin{figure}[htbp]
		\centering
		\includegraphics[width=8cm]{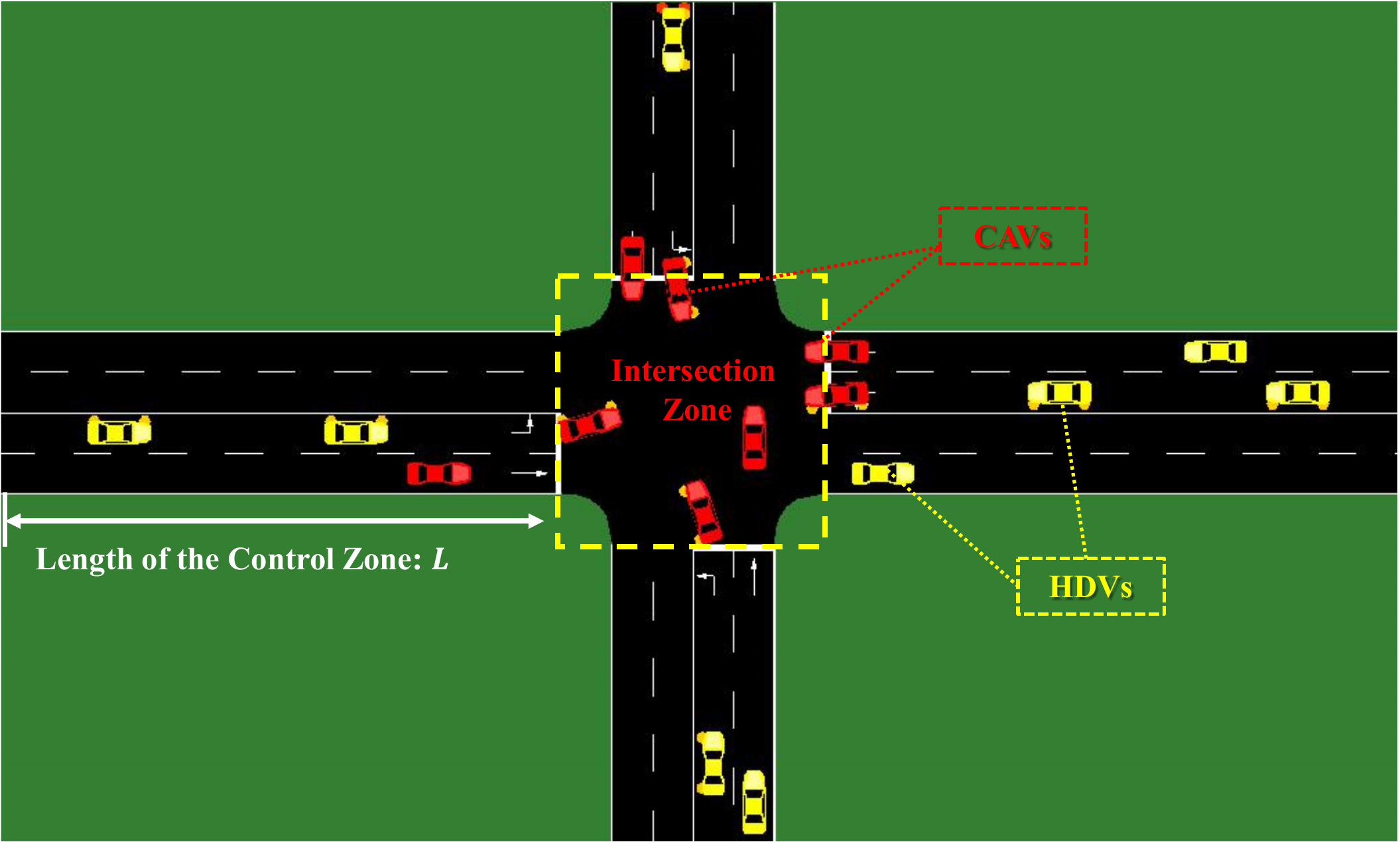}		\caption{Illustration of intersection with four road segments and two lanes. 8 DRL-controlled CAVs are colored in red, and IDM-controlled HDVs in this scenario are colored in yellow. The ranges of intersection and control zone are also shown in this figure.}
		\label{intersection_ill}
	\end{figure}

	\noindent
	We construct an isolated and non-signalized intersection with four ways and two lanes (composed of left-turn and through movements) to solve the CAVs coordination problem under different mixed-autonomy traffic densities of 150veh/hour/lane and 300veh/hour/lane. These density values of traffic flow used in our work are referred to as the parameters of the Grid scenario in research. \cite{vinitsky2018benchmarks}.  Moreover, we employ modified QMIX to coordinate 8 CAVs from each lane. When some of the CAVs leave the intersection, the vehicle behind it will be controlled by the MDRL algorithm, and the rest of the HDVs residing in the control zone will be simulated by the IDM model. The whole simulation is implemented in open-sourced Simulation of Urban Mobility (SUMO) simulator, as illustrated in Fig. \ref{intersection_ill}.
	
    IDM is a typical model aiming to control the mixed traffic at the intersection \cite{RN43} and can accurately represent driver behavior in the real-world \cite{RN162}. This controller sends the acceleration command as follows,
	\begin{equation}
	{a_{IDM}} = a\left[ {1 - {{\left( {\frac{v}{{{v_d}}}} \right)}^\delta } - {{\left( {\frac{{s^*(v,\Delta v)}}{s}} \right)}^2}} \right]
	\label{eq15}
	\end{equation}
	where $a$ is the acceleration term, $v_d$ is the expected speed, $\delta$ serves as an acceleration exponent, the vehicle’s actual headway is denoted by $s$, $\Delta v$ marks the speed difference between the ego and the lead vehicle, and $s^*(\cdot)$ is a function of the speed and the speed difference, which is expressed as:
	\begin{equation}
	s^*(v,\Delta v) = {s_0} + \left( {\max (0,vT + \frac{{v\Delta v}}{{2\sqrt {ab} }})} \right)
	\label{eq16}
	\end{equation}
	
	In which $s_0$ represents the minimum gap, $T$ is a time gap, the comforTable.deceleration is denoted as $\Delta v$. The corresponding parameters adopted in this paper are presented in Table.\ref{tab_IDM}.
	\begin{table}
		\begin{center}
			\caption{IDM parameters}
			\label{tab_IDM}
			\begin{tabular}{c c}
				\toprule
				\textbf{Parameters} &\textbf{Value} \\
				\midrule
				Desired Speed & 15m/s \\
				Time Gap & 1.0s \\
				Minimum Gap & 5.0m \\
				$\delta$ for Acceleration Exponent & 4 \\
				\bottomrule
			\end{tabular}
		\end{center}
	\end{table}
	
	The lateral movements of all cars are fixed by the routes built in SUMO, and the longitudinal motion control is given by:
	\begin{equation}
	\begin{array}{l}
	{x^i}(t + 1) = {x^i}(t) + {v^i}(t) \cdot \Delta  + \frac{1}{2}{a^i}(t) \cdot {\Delta ^2}\\
	{v^i}(t + 1) = {v^i}(t) + {a^i}(t) \cdot \Delta ,
	\end{array}\label{eq6}
	\end{equation}
	where $x^i(t)$ is the travelling distance of vehicle $i$ at time step $t$, $v^i$ and $a^i$ denotes the velocity and acceleration of vehicle $i$ at time step $t$. The discrete time step is $\Delta$. Our approach outputs the acceleration of each vehicle.

	\subsection{RL Formulation}
	\noindent
	The coordination problem has the following assumptions: 
	\begin{itemize}
	    \item[a)] CAVs can communicate perfectly with each other during the training process.
	    \item[b)] Each CAV can only observe its own position, speed. Moreover, to make this application partially observable, CAVs are set not to observe the information of HDVs surrounded.
	    \item[c)] The lateral behaviors of CAV are predefined. 
	    \item[d)] The behaviors of HDVs are controlled by IDM. 
	\end{itemize}
    The main goal of our distributed controller is to output the longitudinal acceleration for each vehicle. Depending on those assumptions, the transform to RL problem is shown as follows.
	\subsubsection{State, observation and action space}
	\noindent
	The observation space of each CAV contains the following information (the notations are consistent with that in the Dec-POMDP definition):
	\begin{equation}
	{o^i} = [x_i,y_i,v_i]
	\label{eq7}
	\end{equation}
	in which $[x_i,y_i,v_i]$ presents the individual vehicle’s ($i=1,2,\dots,8$) position in Cartesian coordinated originated at the intersection center and speed respectively. Consequently, the global state space is simply the concatenation of all individual’s observations, i.e., $s = [{o^i}]_{i = 1}^N$, where $N$ is the number of controlled CAVs.
	
	As for the actions taken by the agents, some RL research endeavors to deal with the continuous action to control real-world robotic tasks \cite{lillicrap2016continuous,RN52}. However, we construct a discrete action space for the purpose of lower computational demand. It is natural to extend the dimensions of the discrete action space for controlling the agents at a smaller granularity level when applied to real-world scenarios. The definition of action and joint action space is shown as follows, the individual action space is ${a^i} = [[a_j]_{j=1}^{p},0,[d_j]_{j=1}^{p}]$ coded in a one-hot manner, where $p$ denotes the granularity of discretization. Only one element can be selected at each time step. For example, an action may be $[1,0,0,0,0,0,0]$ (where $p=3$), meaning that the agent $i$ will accelerate with $a_1$. The acceleration and deceleration is clamped not to surpass the corresponding extremums. The joint action space is the concatenation of individual action, $[a^i]_{i=1}^N$.To facilitate the policy training, we collect the history action-observation trajectories as $\tau^i=(o^i\times a^i)$ for each agent. Moreover, the action masking mechanism similar to the work \cite{samvelyan2019starcraft} is incorporated into the design of valid action spaces of agents. Specifically, when the vehicle approaches the other within the minimal head-to-tail distance, it must slow down with available discrete deceleration values.
	
	\subsubsection{Reward setting}
	\noindent
	Designing the reward function in the RL algorithm is of paramount significance, implicitly influencing the convergence speed and asymptotic performance of the algorithm. Empirically, more negative rewards can stimulate the agent to finish the episode with minimal timesteps due to cumulative punishments. On the other hand, more positive rewards motivate the agents to gather more prizes as they can, hence consuming more timesteps. 
	
    In our scenario, the vehicles need maximize their speed, and simultaneously guarantee safety within an episode. As a result, our reward function is composed of three macro perspectives: 1) traffic efficiency; 2) collision avoidance (for safety);
	
	As for the efficiency, the low-speed and the long waiting time at the control and intersection zone can be penalized, and the higher speed than the predefined threshold is encouraged. The expression is shown as:
	\begin{equation}
	{r_{eff}} =  - \mathop \sum \limits_{i = 1}^N \alpha _1 \cdot\mathbb{I}({v_i} < {V_{min}}) +{\sum_{i=1}^N\alpha_2 \cdot v_i}
    \label{eq8}
	\end{equation}
	in which $\mathbb{I}(\cdot)$ is the indicator function denoting that the function value is set to 1 if ${v_i} < {V_{min}}$ (where $v_i$ is the normalized speed ranging from $0$ to $1$, $V_{min}$ is the minimal allowed speed for vehicles in this scenario), or 0 otherwise. The rest of parameters: $\alpha_1, \alpha_2$ are adjustable during the training. 
	
	The safety of each vehicle can be guaranteed by the setting of the reward function such that the collision cases can be severely punished. This part is formulated as follows:
	\begin{equation}
	{r_{ca}} =  - \mathop \sum \limits_{i = 1}^N \;{\alpha _3}\cdot\mathbb{I}({\rm{Collision}})
	\label{eq9}
	\end{equation}
	where the collision is detected by the relative distance of vehicles. Compared to enumerating all conflict points at the intersection, this setup can reduce the computational complexity and ease the implementation in practice. $\alpha_3$ is also a variable parameter during the training.  
	
	Besides, in our scenario, each episode with a fixed length will record the number of collision times where the collision criterion is that the relative distance of two vehicles (head-to-tail distance) is smaller than the predefined threshold shown in Table.\ref{simParams}
	
        \section{Methods}
	\label{methods}
	\subsection{Architecture of QMIX Algorithm}
\begin{figure}
		\centering
		\includegraphics[width=8.7cm,height=9.04cm]{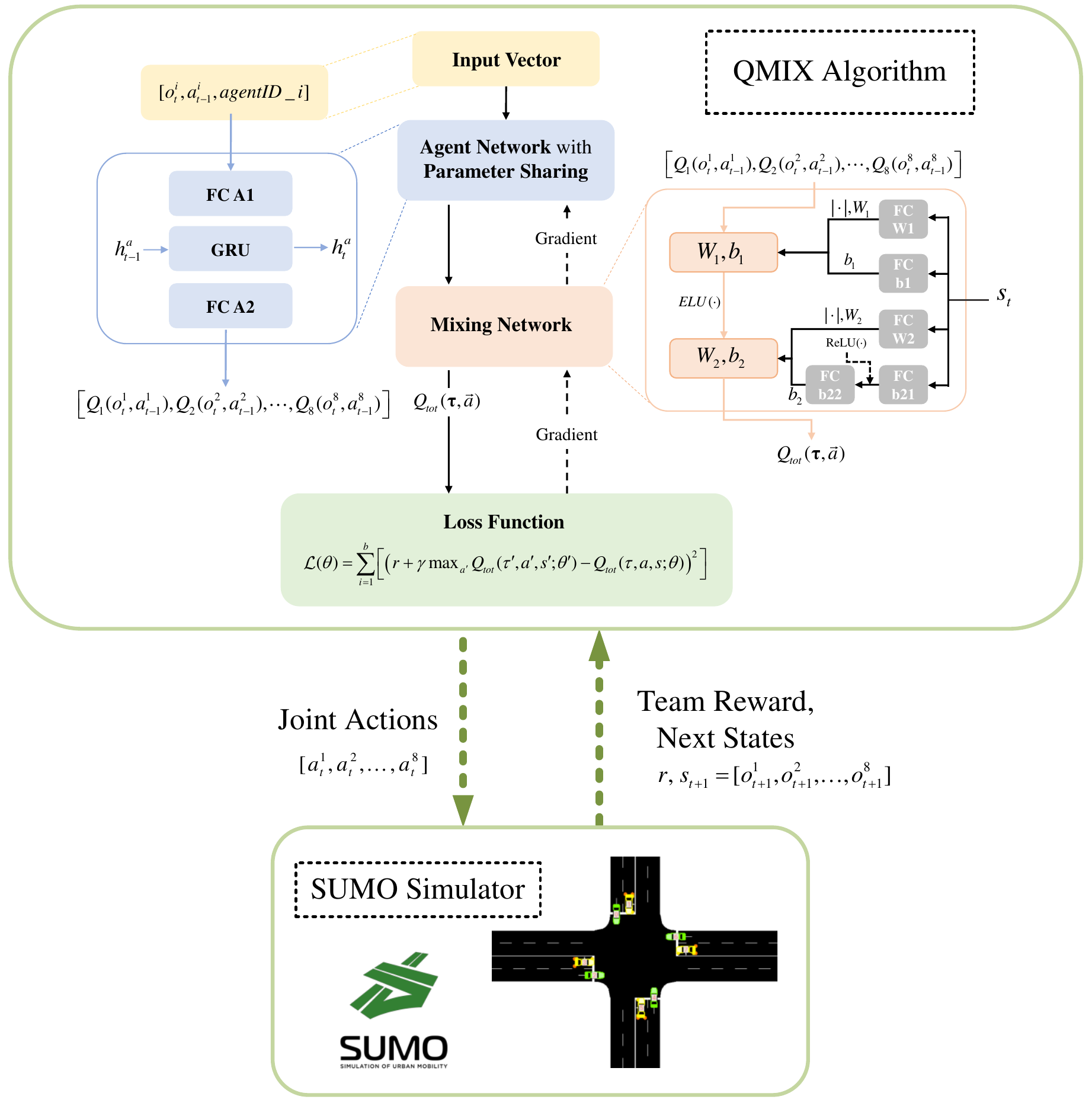}
		\caption{The overall architecture of QMIX algorithm with specified input and output using the related definitions of intersection scenario, where FC is short for fully connected layer, the alphanumeric description after FC aims to distinguish each FC layer and GRU for the gated recurrent network.}
		\label{algo_framework}
	\end{figure}

	\noindent 
	The framework of QMIX with a detailed description of input and output is shown in Fig. \ref{algo_framework}. The agent network and mixing network are two pedestals of this approach from the forward perspective and all controlled agents share the same networks (i.e., parameter sharing). The merit of parameter sharing is that it can considerably avoid the scalability issue as the dimension of action space does not explode with the increasing number of learning agents. The input to the agent network is the local observation vector of each vehicle plus its identification vector coded in one-hot form. It outputs the estimate of action value $Q_i$ of the agent $i$ which is fed into the mixing network. The mixing network then takes as input the global state of the environment plus the estimate of actions values. It utilizes the hypernetworks composed of several fully connected network layers with different activation functions such as rectified linear unit (ReLU) units \cite{RN153} to compute the weights and biases ${W_1},{W_2},{b_1},{b_2}$. One further significant remark is that the hypernetwork uses an absolution operation to generate weights ${W_1},{W_2}$ such that the monotonicity constraint can be properly enforced and hence makes it a sufficient condition for IGM principle. 
	
	The eventual output of the mixing network is $Q_{tot}$, a rough estimate of the immediate team reward after the joint actions are taken by agents. The parameters of all networks are updated by the gradient descent algorithm by computing the gradient of the loss function. The full implementation of QMIX is based on Pytorch \cite{RN155}, a machine learning library. During the training, each agent adopts the $\epsilon$-greedy policy ($\epsilon$ controls the balance of exploration and exploitation of each agent, decreasing with the rising number of training steps) to choose the action.
	
	\subsection{Implementation Tricks}
	\noindent
	Recent work demonstrates that improving the code-level implementations of QMIX can positively affect its asymptotic performance \cite{RN54}. Therefore, we are inspired to perform the following tricks:

	1) The eligibility traces based on Peng’s Q($\lambda$) \cite{RN156} method is exploited in our work. This trick empirically enables QMIX to become stable and converge faster and avoids the significant bias against bootstrap returns caused by non-sufficiently trained Q networks \cite{RN54}. The formula of Peng’s Q($\lambda$) for the mixing network is shown below:
	\begin{equation}
	\begin{array}{l}
	G_l^\lambda \dot  = (1 - \lambda )\sum\limits_{n = 1}^\infty  \; {\lambda ^{n - 1}}{G_{l:l + n}}\\
	{G_{l:l + n}}\dot  = \mathop \sum \limits_{t = l}^{l + n} \;{\gamma ^{t - l}}{r_t} + {\gamma _{n + 1}}\mathop {\max }\limits_{\bm a} {Q_{tot}}({\bm{\tau }},\bm a,{s_t};{\theta _{tot}})
	\end{array}
	\label{eq13}
	\end{equation}
	in which $G_l^\lambda$ replaces the $y^{\rm{target}}$ in eq. (\ref{eq3}), and $\lambda$ is the discount factor of traces with the property that $\prod _{l = 1}^t\lambda=1$ when $t = 0$.
	
	2) Clipping the reward at each time step to a fixed range can stabilize the training for deep Q learning \cite{mnih2015human}. We regard this range as an adjusTable.parameter, and the value is shown in Table. \ref{RL_env_params}. 
	
	3) Network-level improvements: after tweaking the components at the network level, we find out that it is useful to alter the optimizer of the QMIX from RMSProp \cite{RN189} to Adam \cite{kingma2015adam}  with exponentially decaying learning rate \cite{exp_lr2013}, improve the initialization of network parameters from a uniform distribution to the Xavier normal distribution \cite{RN159}, and employ the orthogonal initialization for GRU. The experimental results in section \ref{results} show the improved efficacy of QMIX with our implementation tricks compared to the original one.
	
	\section{Simulation and Analysis}
	\label{results}
	\subsection{Experimental Settings}
	\noindent
	For clarity, Fig. \ref{intersection_ill} illustrates the DRL-controlled CAVs colored in red and IDM-controlled HDVs colored in yellow, and Table.\ref{simParams} shows the parameters of our simulation environment established on SUMO 1.10.0. To ease the implementation and environment construction, we set the lane length of each road segment to 100m, and vehicles in traffic flow with different density values (150 and 300veh/hour/lane) approach the intersection at random speeds. Moreover, the absolution value of discretization granularity of acceleration and deceleration is symmetric, $[1.5,2.5,3.5]\ m/s^2$. 
	
	Table.\ref{RL_env_params}  summarizes all parameters related to the RL problem, specifying the dimensions of state, action space, and maximum steps of one episode. 
	
	To the best of our knowledge, PPO is the most commonly used DRL algorithm applied to the non-signalized intersection \cite{RN33, RN43, 2021Connected, vinitsky2018benchmarks}. The parameters of the PPO we implement are shown in Table. \ref{PPO_params}. Hence, we first train two  MDRL algorithms, including the original and modified QMIX and one single-agent DRL algorithm PPO, in the scenario with the traffic flow of the density 150veh/hour/lane. Afterward, the density of traffic flow is doubled to 300veh/hour/lane, and on which the modified QMIX policy is trained. Furthermore, all training processes are conducted under three random seeds to exclude the influence of randomness. During the training phase, the policy with the maximum reward and the minimum number of collisions is stored and evaluated by average speed and fuel consumption metrics. Besides, the hyperparameters of the QMIX algorithm and some tricks-related parameters throughout the training stage are jointly summarized in Table.\ref{hyperparams}. All simulations were conducted on a single computer with 3.7GHz Intel(R) Core(TM) i7-8700K CPU and NVIDIA GeForce RTX 2080 GPU.
	
	\begin{table}
		\begin{center}
			\caption{Simulation parameters.}
			\label{simParams}
			\begin{tabular}{c c}
				\toprule
				\textbf{Parameters} &\textbf{Value} \\
				\midrule
				Lane Length & 100m\\
				Lane Width & 3.2m\\
				Length of CAV & 5m \\
				Maximal Allowed Speed of CAV & 15m/s \\
				Initial Speed of All CAVs & random \\
				Minimal Allowed Speed of CAV & 2m/s \\
				Acceleration Granularity & [1.5,2.5,3.5]m/s$^2$ \\
				Deceleration Granularity & [-1.5,-2.5,-3.5]m/s$^2$ \\
				Safe Relative Distance at Intersection & 0.2m \\
				SUMO Version & 1.10.0 \\
				\bottomrule
			\end{tabular}
		\end{center}
	\end{table}
	
	\begin{table}
		\begin{center}
			\caption{RL environment parameters}
			\label{RL_env_params}
			\begin{tabular}{c c}
				\toprule
				\textbf{Parameters} &\textbf{Value} \\
				\midrule
				The Number of Controlled CAVs & 8 \\
				Discrete Time Step & 0.1s \\
				Episode Timesteps & 200 \\
				Dimension of Individual Observation Space & 10 \\
				Dimension of Individual Action Space & 7 \\
				Dimension of Global State Space & 80 \\
				Maximum Reward at Each Step & 10 \\
				Minimum Reward at Each Step & -5 \\
				\bottomrule
			\end{tabular}
		\end{center}
	\end{table}

	\begin{table}
		\begin{center}
			\caption{Hyperparameters in QMIX training}
			\label{hyperparams}
			\begin{tabular}{c c}
				\toprule
				\textbf{Parameters} &\textbf{Value)} \\
				\midrule
				FC\_A1 & [16, 64] \\
				GRU & [64, 64] \\
				FC\_A2 & [64, 3] \\
				FC\_W1 & [64, 256] \\
				FC\_b1 & [64, 32] \\
				FC\_W2 & [64, 32] \\
				FC\_b21 & [64, 32] \\
				FC\_b22 & [32, 1] \\
			    Total Training Timesteps & 1500000 \\
		    	Optimizer & Adam \\
			    Learning Rate & $1\times 10^{-4}\to 0$ \\
			    Initial Value of  $\epsilon$ in $\epsilon$-greedy Policy & 1.0 \\
    			Terminate Value of $\epsilon$ & 0.05 \\
    			Annealing Steps for $\epsilon$ & 100000 \\
    			Batch Size & 64 \\
    			$\lambda$ in Q($\lambda$) & 0.4 \\
    			Target Update Cycle & 100 \\
    			learning rate decay & 0.991 \\
    			$\alpha_1$ & 0.5 \\
    			$\alpha_2$ & 1.0 \\
    			$\alpha_3$ & 5 \\
				\bottomrule
			\end{tabular}
		\end{center}
	\end{table}
	
	\begin{table}
		\begin{center}
			\caption{Parameters of PPO}
			\label{PPO_params}
			\begin{tabular}{c c}
				\toprule
				\textbf{Parameters} &\textbf{Value} \\
				\midrule
				Discount factor $\gamma$ & 0.99 \\
				$\lambda$ & 0.95 \\
				Clip range $\epsilon$ & 0.2 \\
				Total timesteps & 1500000 \\
				Minibatch size & 8 \\
				Epoch & 4 \\
				Learning rate& 0.0003 $\to$ 0 \\
				Hidden layer number & 2 \\
				Hidden units number & 128 \\
				Initialization & orthogonal\\
				Optimizer & Adam \\
				\bottomrule
			\end{tabular}
		\end{center}
	\end{table}
		
	\subsection{Comparison of QMIX policy and PPO under a fixed traffic density}
	\noindent
	
\begin{figure}
		\centering
		\captionsetup{justification=centering}
		\subfloat[Episode Reward versus Episodes]{\label{eps_rew}\includegraphics[width=0.46\textwidth]{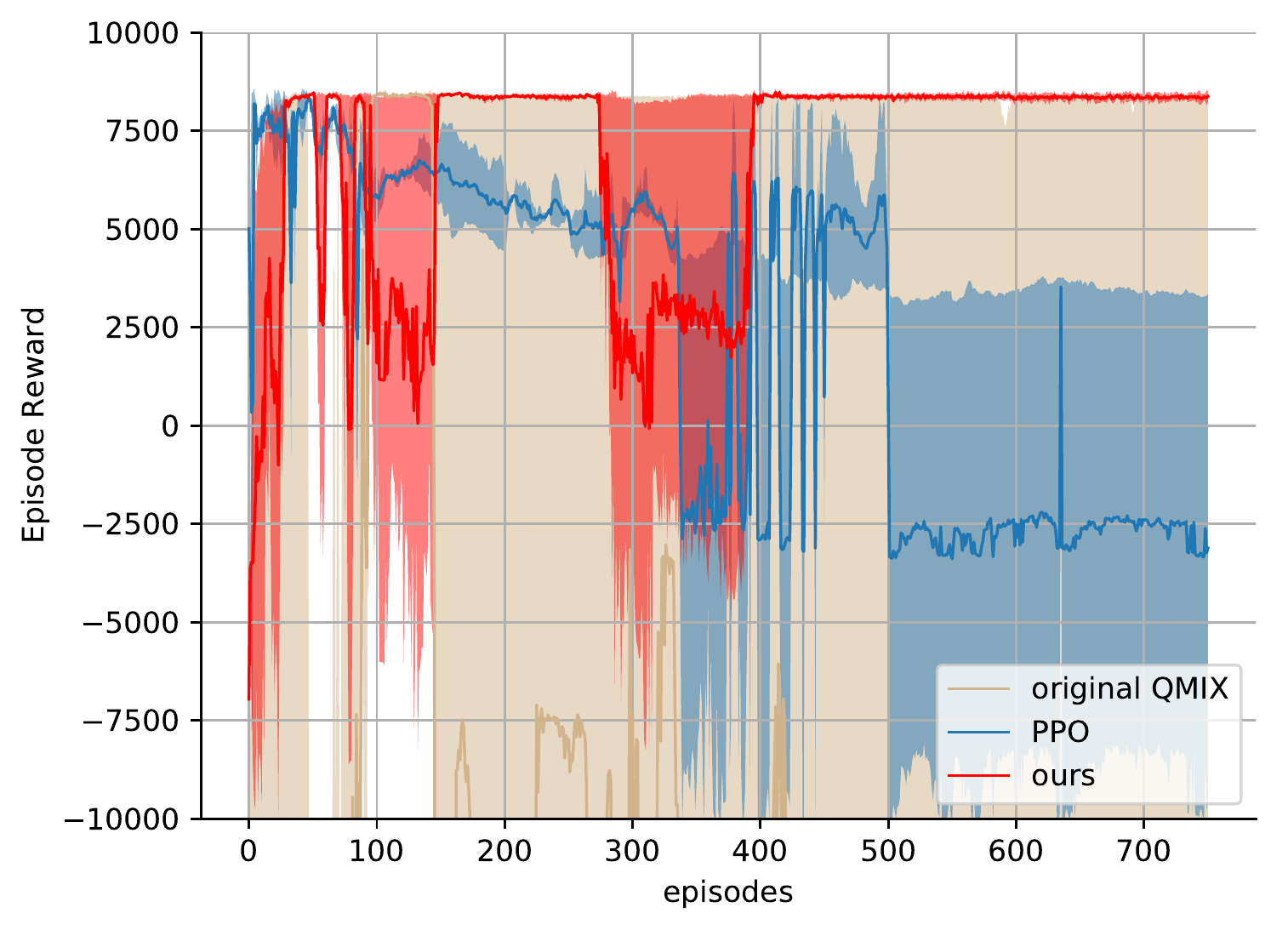}}\\
		\subfloat[Collision Times vs Episodes]{\label{coll_times}\includegraphics[width=0.46\textwidth]{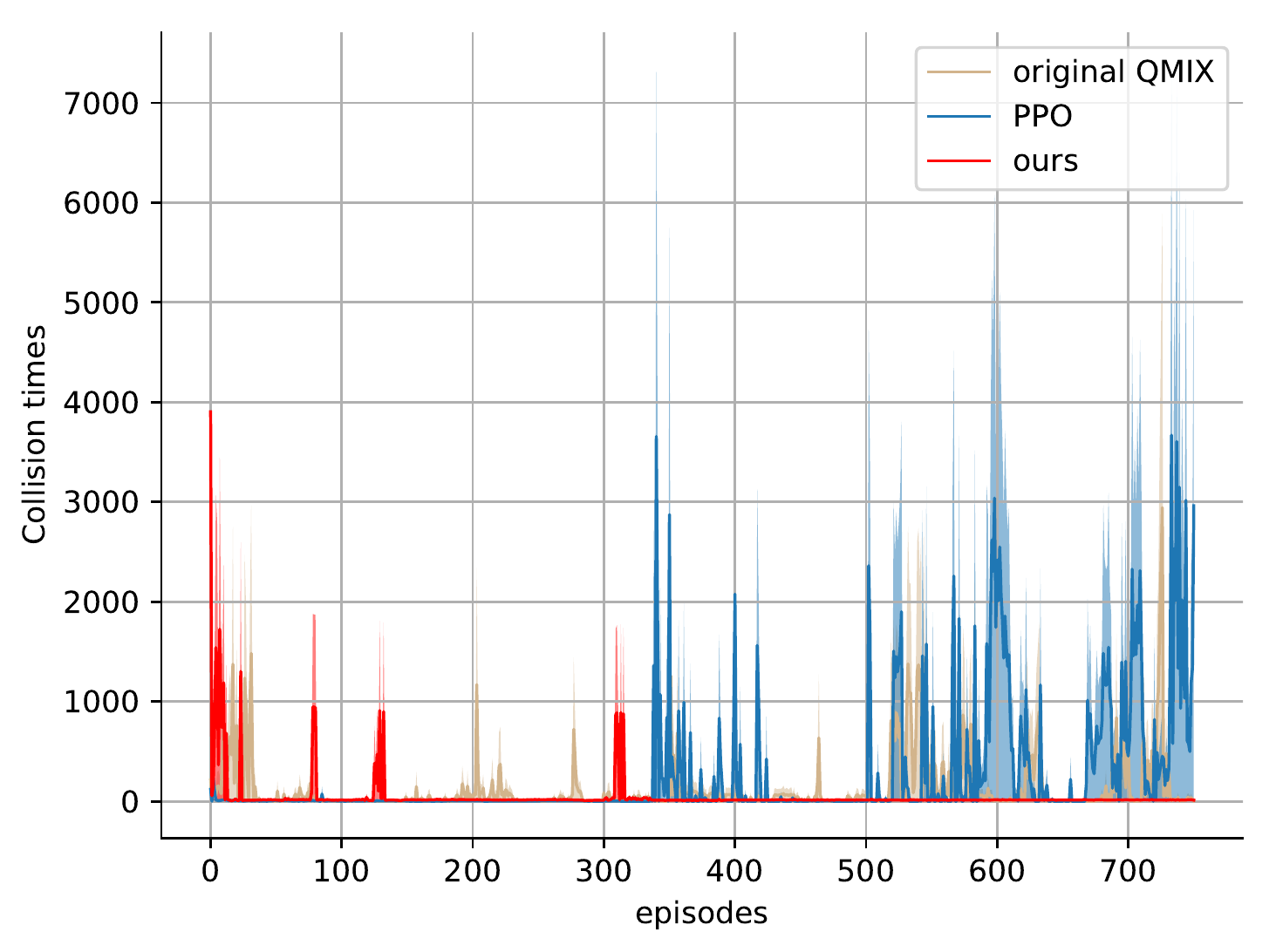}}\\
		\caption{The training process of the modified QMIX, the original QMIX and PPO. including (a) the change of episode reward, and (b) records of the change of the number of collisions per episode. The experiments are conducted under three random seeds.}
		\label{training}
	\end{figure}

	We first compare the learning performance comprising the change of episode reward and the number of collisions in one episode, as shown in Fig. \ref{training}, in which the maximal and minimal values determine the shaded area implying the variance in each subplot. In Fig. \ref{eps_rew}, the episode reward of the modified QMIX first climbs quickly to nearly the maximum value under the effect of Xavier normal and orthogonal initialization schemes introduced in section \ref{methods}. As the progress of active exploration, agents may first fall into poor decisions, in which the episode reward collapses. Then they learn from their erratic experiences and gradually find the correct action, as they change the strategy from exploration to exploitation. In the end, the episode reward for the modified QMIX converges with the minimum variance compared to the other two methods. The variance of episode reward for the original QMIX is monstrous, which may be caused by the lack of reward clipping, the absence of eligibility traces, and network-level improvements. Moreover, it does not converge, and the mean of its asymptotic performance is significantly lower than the modified counterpart.\\
	
    Moreover, although the learning performance of PPO stays superior at the beginning of the training, it becomes weakened with the exploration progress, and its final mean of the episode reward declines and becomes lower than that of the modified QMIX with more variance. In Fig. \ref{coll_times}, the number of collisions of modified QMIX finally converges to nearly zero with slight fluctuation. Unsurprisingly, the values in the curves of the other two intend to diverge as the progress of training, which raises the safety issue and indicates the divergence of these two algorithms.
	
\begin{figure}[!t]
		\centering
		\subfloat[At Timestep 3.7s]{
			\includegraphics[width=0.23\textwidth]{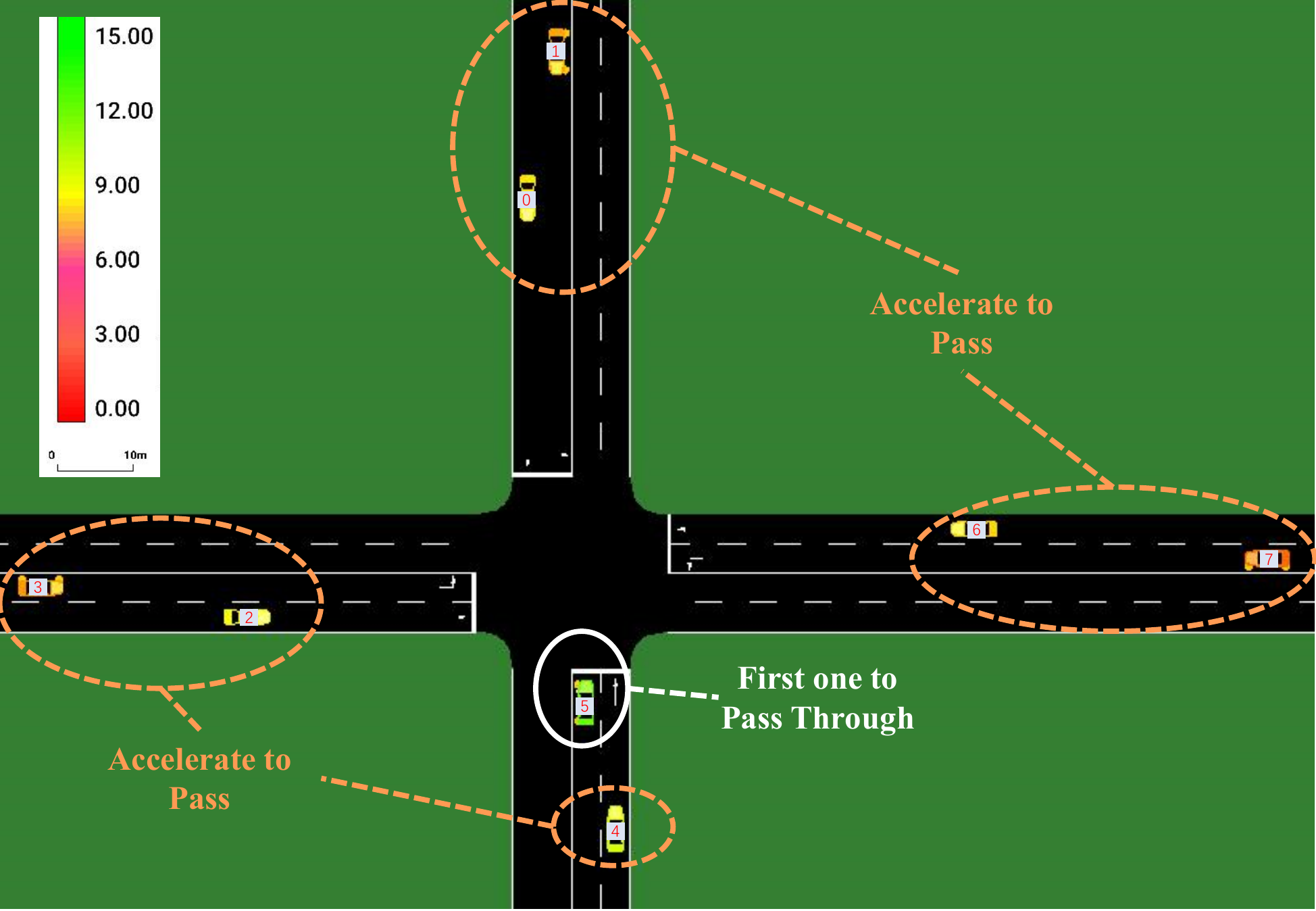}
			\label{ill_1}}
		\subfloat[At Timestep 6.5s]{
			\includegraphics[width=0.23\textwidth]{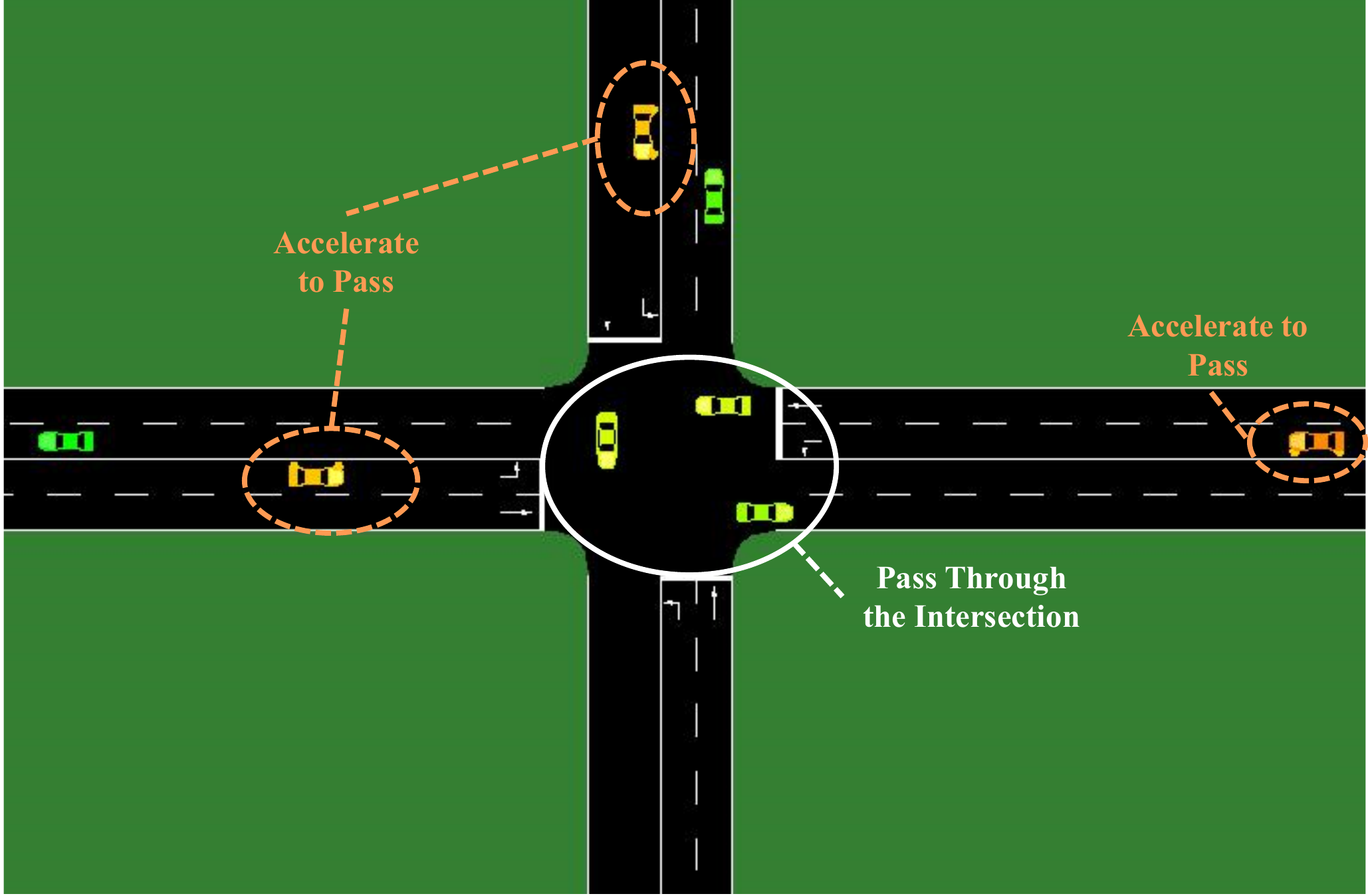}
			\label{ill_2}}\\
		\subfloat[At Timestep 9.2s]{
			\includegraphics[width=0.23\textwidth]{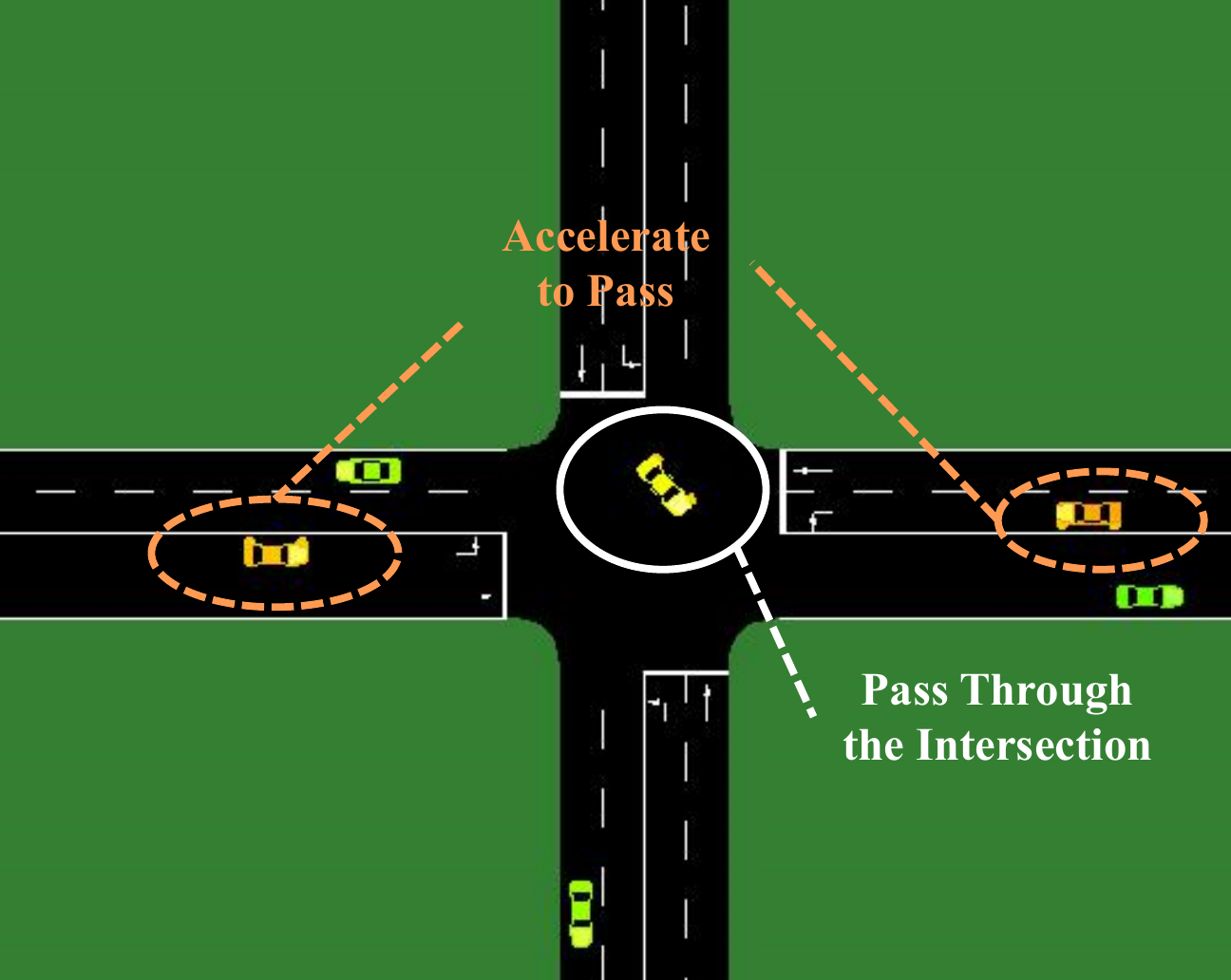}
			\label{ill_3}}
		\subfloat[At Timestep 12.1s]{
			\includegraphics[width=0.23\textwidth]{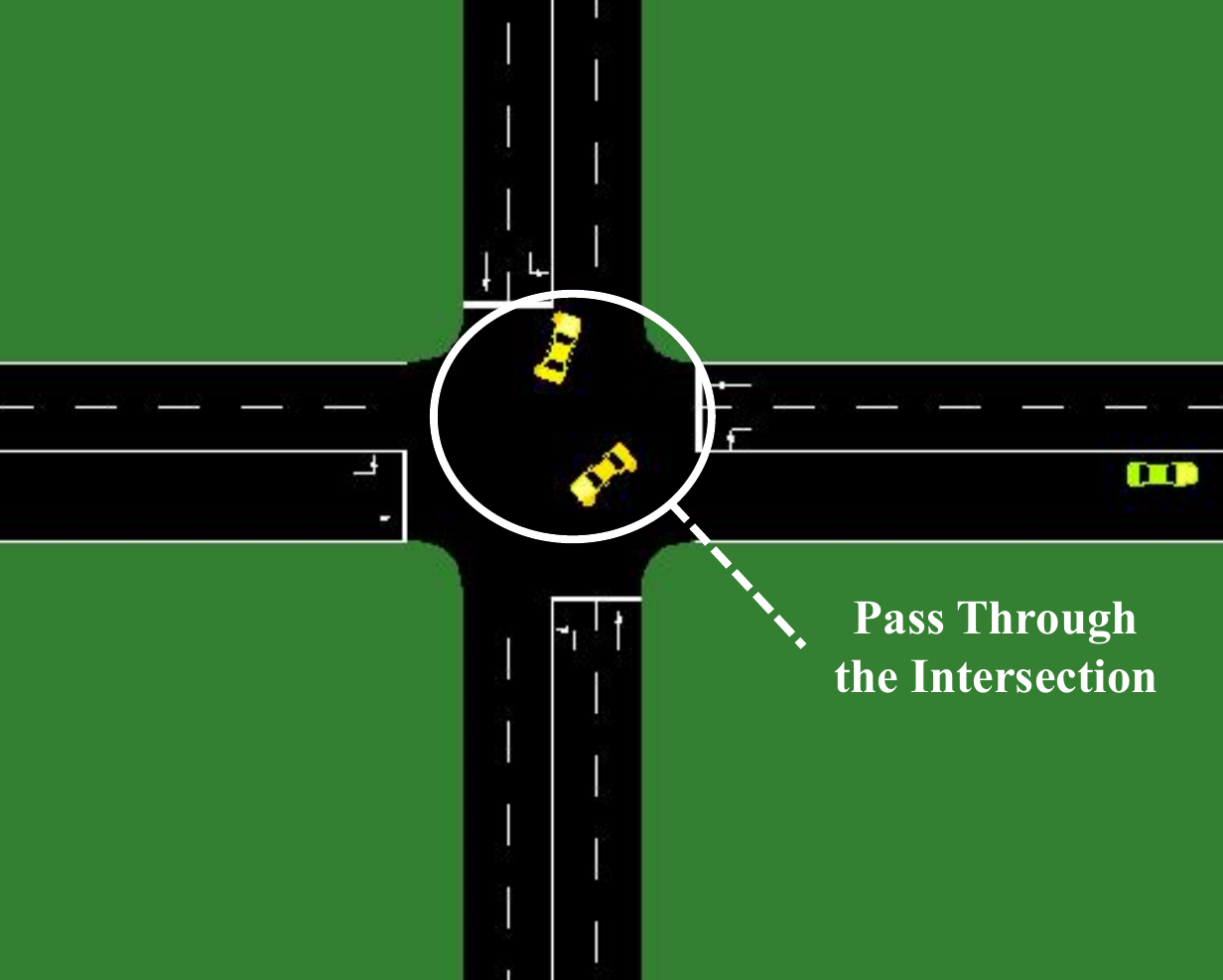}
			\label{ill_4}}\\
			
		\caption{A set of snapshots produced by SUMO-GUI at different timesteps with several illustrative marks aims to demonstrate the efficacy of the modified QMIX policy to coordinate vehicles at a non-signalized intersection under the traffic flow of 150veh/hour/lane. The color of each vehicle marks its current speed.}
		\label{snapshots}
	\end{figure}

	During the training process, we store the model of each DRL or MDRL algorithm with the highest reward and the lowest number of crashes. Moreover, we take the snapshots of the trained QMIX policy with implementation tricks to intuitively illustrate its emergent behaviors under the traffic flow of 150veh/hour/lane, as shown in Fig. \ref{snapshots}. Since this traffic density is relatively low, each road segment for a particular time duration only contains CAVs, and after they pass through the intersection safely, a time period in which there are no vehicles exists. At 3.7s, CAV 5 enters the junction with a maximal speed of 15m/s, and the rest of the vehicles on the road attempt to accelerate. Afterward, at 6.5s, vehicle 4 leaves the intersection zone with maximum speed, and vehicles 0, 2, and 6 speed up to enter the junction. Only one vehicle (with index 1) enters the intersection at 9.2s. Finally, vehicles 3 and 7 enter the zone, and the others accelerate to maximum speed to leave the control zone.\\
	
\begin{figure}[!t]
		\centering
		\subfloat[Average Speed vs Timestep]{\includegraphics[width=0.46\textwidth]{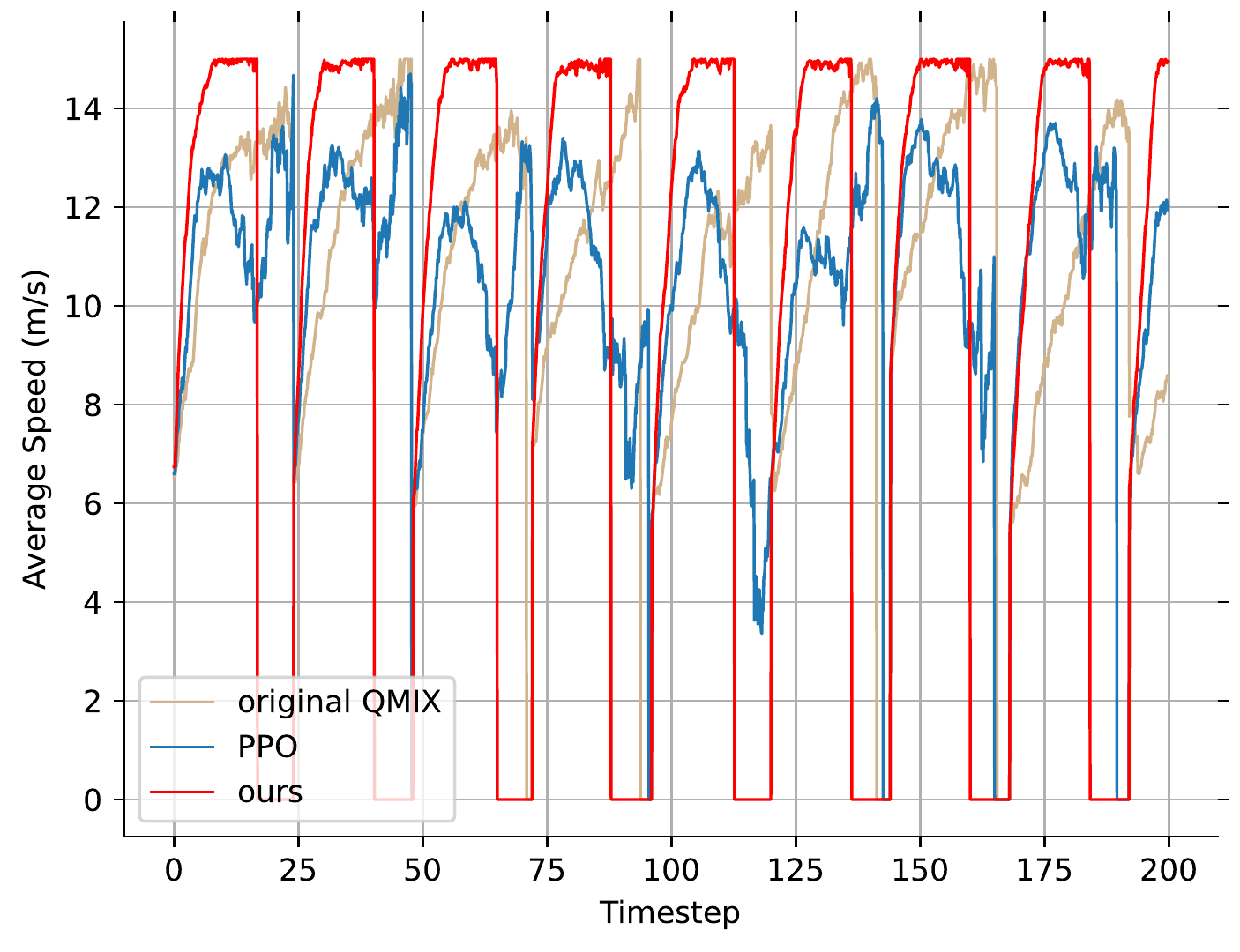}\label{vel_compare}}\\
		\subfloat[Average Fuel Consumption vs Timestep]{\includegraphics[width=0.46\textwidth]{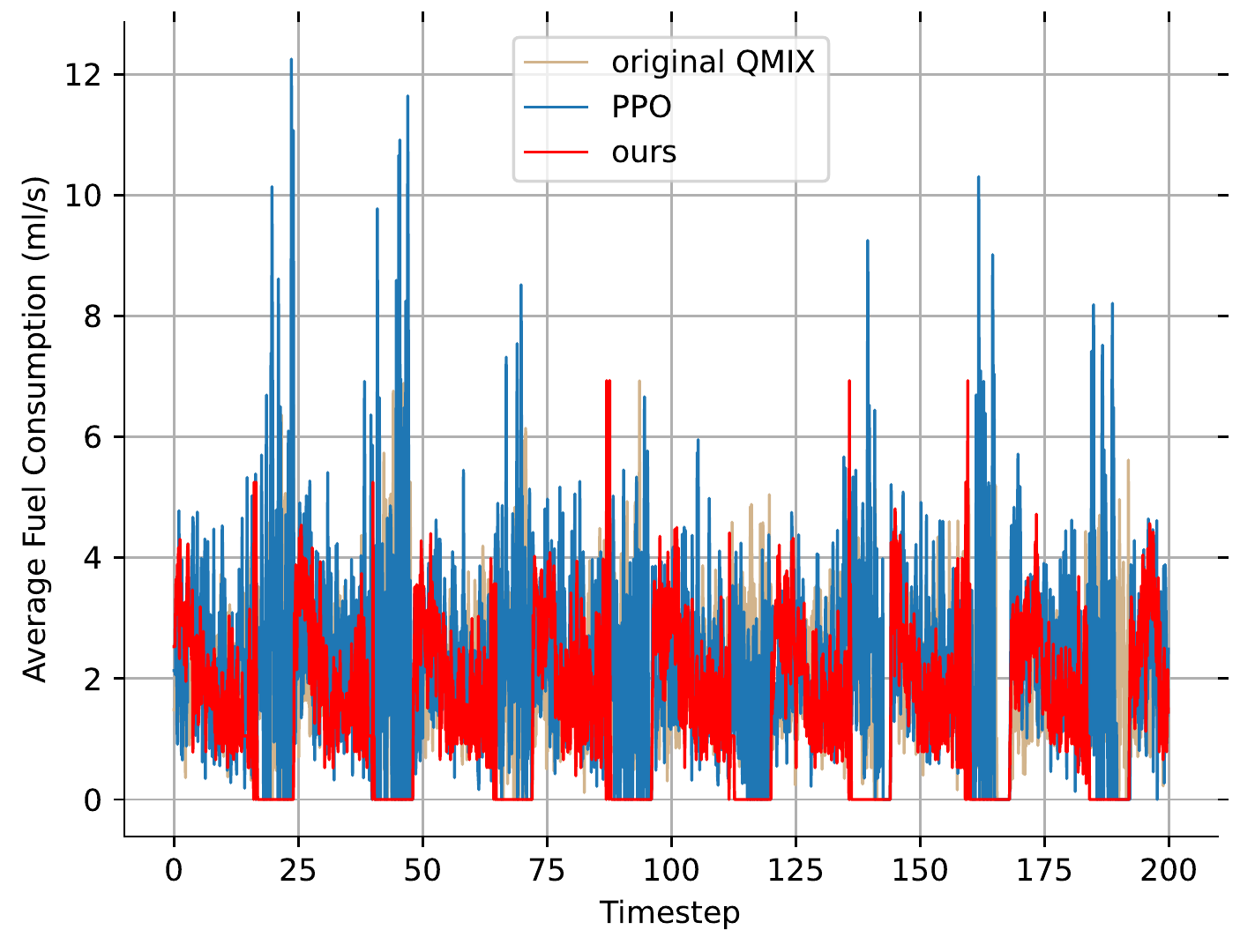}\label{fuel_compare}}\\
		\caption{(a) The average speed of all vehicles on the road, where CAVs are controlled by the stored policies (the modified QMIX, the original QMIX, and PPO) in one episode. (b) The average fuel consumption of all vehicles on the road, in which CAVs are controlled by stored policies (the modified QMIX, the original QMIX, and PPO) in one episode.}
		\label{comparsion_metrics_150}
	\end{figure}

	Then we evaluate the stored models of these three algorithms under three metrics: (1) The change of average speed of all vehicles in the scenario with timestep in one episode. (2) The change of average fuel consumption of all vehicles with timestep per episode. (3) The number of collisions per episode for these three algorithms. These experimental results are recorded as Fig. \ref{comparsion_metrics_150} and Table. \ref{metrics_compare_tab}. In Fig. \ref{vel_compare}, the speed pattern of the modified QMIX model first moves up from the initial speed value and swiftly reaches nearly the maximum speed. The climbing phase and plateau of speed correspond to the time duration when vehicles enter and leave the intersection and finally leave the control zone, respectively. The zero average speed afterward in this curve represents a period where no vehicle is on the road since one group of CAVs controlled by QMIX efficiently leaves the intersection, and the next group is awaiting to be generated by the SUMO simulator. The longer this period becomes, the more efficient the traffic can be under lower traffic density. This phenomenon will disappear when the traffic density increases, as shown in the next subsection. The line graph of fuel consumption follows an approximate opposite trend with some spikes as shown in \ref{fuel_compare}. The average fuel consumed by all vehicles on the road decrease with fluctuations as they speed up, and the spikes appear when vehicles drive to their maximum speed or constantly change their decisions of accelerating or slowing down.
	
	However, the other two methods' speed and fuel consumption patterns are not as clear as the modified QMIX. Concerning the speed profile of vehicles controlled by the pure QMIX model, the speed-increasing rate and the duration of zero speed are lower than the modified one. Vehicles frequently change their speed, leading to variations during the climbing phase. The behavior of more frequent alternation between the acceleration and deceleration emerges in the baseline case (PPO), causing the traffic inefficiency and more consumption of fuels shown in \ref{fuel_compare}. As summarized in Table.\ref{metrics_compare_tab}, vehicles controlled by the modified QMIX can drive with the most significant average speed at 13.52m/s and consume the least average fuel of 2.06ml/s among these three approaches. Noticeably, these values are computed by wiping out the zero-speed duration. Moreover, these two QMIX methods can eliminate the phenomenon of vehicle crashes, but PPO cannot.
	
	\begin{table}
		\begin{center}
			\caption{Comparison of QMIX and PPO}
			\label{metrics_compare_tab}
			\begin{tabular}{c c c c}
				\toprule
				 & \textbf{Ours} & \textbf{Original QMIX} & \textbf{PPO} \\
				\midrule
				Average speed & \textbf{13.52m/s} & 11.32m/s & 10.48m/s \\
				Average fuel consumption & \textbf{2.06ml/s} & 2.20ml/s & 2.27ml/s  \\
				Collisions & \textbf{0} & \textbf{0} & 4 \\
				\bottomrule
			\end{tabular}
		\end{center}
	\end{table}

	\subsection{Comparison of results on different scenario cases}
	\noindent
	
    In this part, we double traffic density to 300veh/hour/lane and retrain our modified QMIX algorithm for this scenario. The influence of more vehicles on the training process is first demonstrated, where the randomness of learning is eliminated by training on three random seeds, similar to the previous subsection. Then we evaluate the average speed, fuel consumption, and the number of collisions of the stored model with the maximum reward and the minimum crashes during the training stage and compare it with its counterpart in the 150veh/hour/lane scenario.

	\begin{figure}[!t]
		\centering
		\subfloat[Episode Reward versus Episodes]{\includegraphics[width=0.46\textwidth]{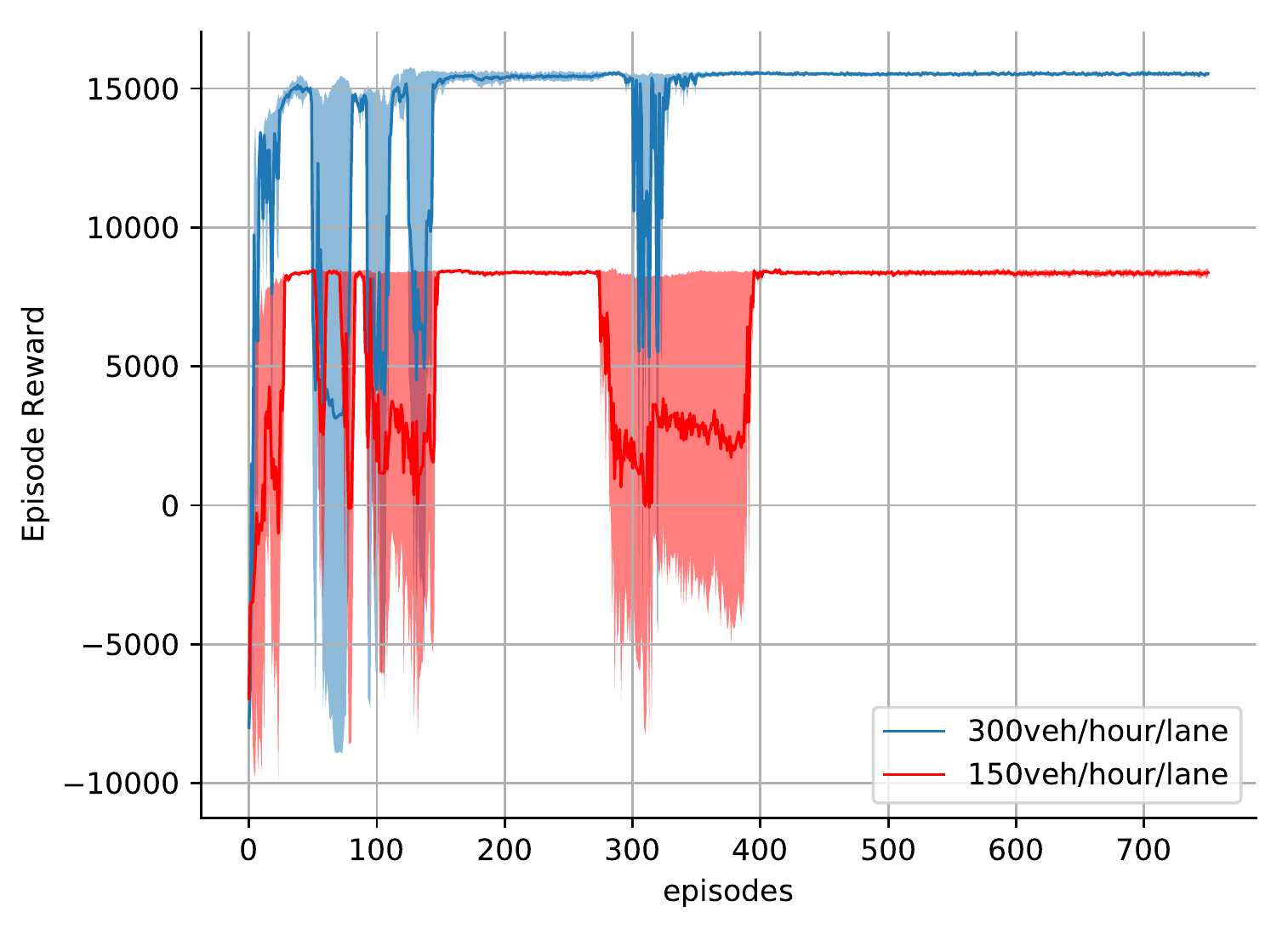}\label{rew_diff_den}}\\
		
		\subfloat[Collisions Times versus Episodes]{\includegraphics[width=0.46\textwidth]{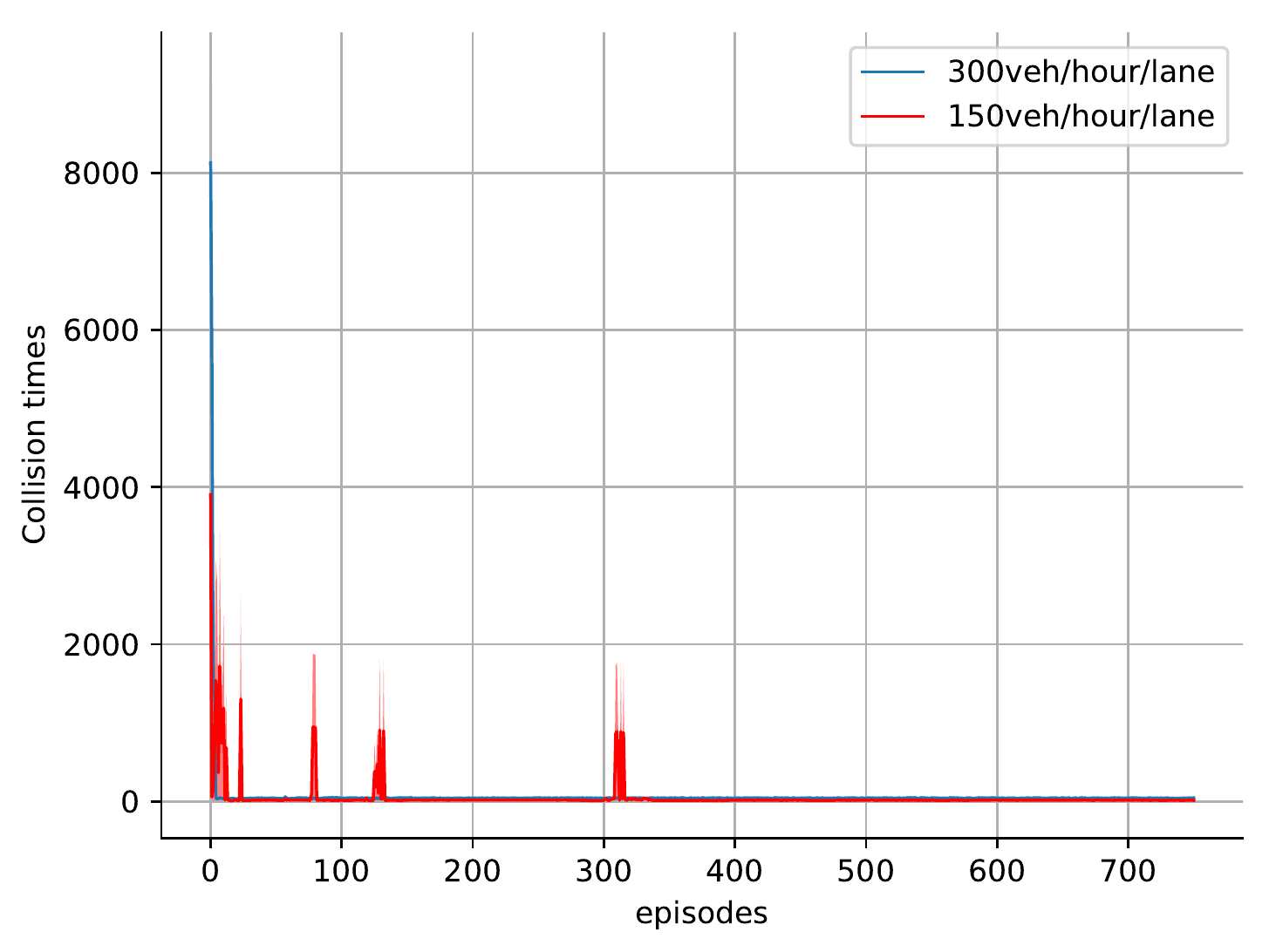}\label{col_diff_den}}\\
		\caption{The training process of applying the modified QMIX algorithm to different traffic densities with values 150veh/hour/lane and 300veh/hour/lane under three random seeds. (a) The change of episode reward under two different traffic densities, and (b) the change of the number of collisions under two different traffic densities.}
		\label{training_diff_den}
	\end{figure}
	
	Some insights can be obtained from the learning curves in Fig. \ref{training_diff_den} under different traffic densities. In general, the reward-learning processes under both scenarios (Fig. \ref{rew_diff_den}) eventually converge. Both demonstrate substantial variances (as shown in the shaded area) during the early heavy exploration period. When agents explore the environment, the reward signal may dramatically fall due to the wrong decision sequences of different acceleration or deceleration values, despite the problem formulation containing the action masking mechanism described in the previous section. Seemingly, the task of coordinating fewer vehicles on the road may cripple policy training. The reason may be that this circumstance brings more local optima to the learning process and confuses the algorithm not to find the global optimum, obtaining the maximal reward and lowest collisions. Further, the reward under the scenario with more dense traffic asymptotically exceeds the counterpart of lower traffic flow since more vehicles with high speed can accumulate more rewards during one episode (Eq. \ref{eq8}). The convergence of our modified QMIX applied to both scenarios validates its efficacy. In Fig. \ref{col_diff_den}, at the beginning, the car crashes is substantially higher than the period afterward. Although these values under both scenarios tend to converge as the learning rewards, the scenario of more vehicles on the road eventually demonstrates slightly more crashes than the lower traffic scene.
	
\begin{figure}[!t]
		\centering
		\subfloat[]{\includegraphics[width=0.46\textwidth]{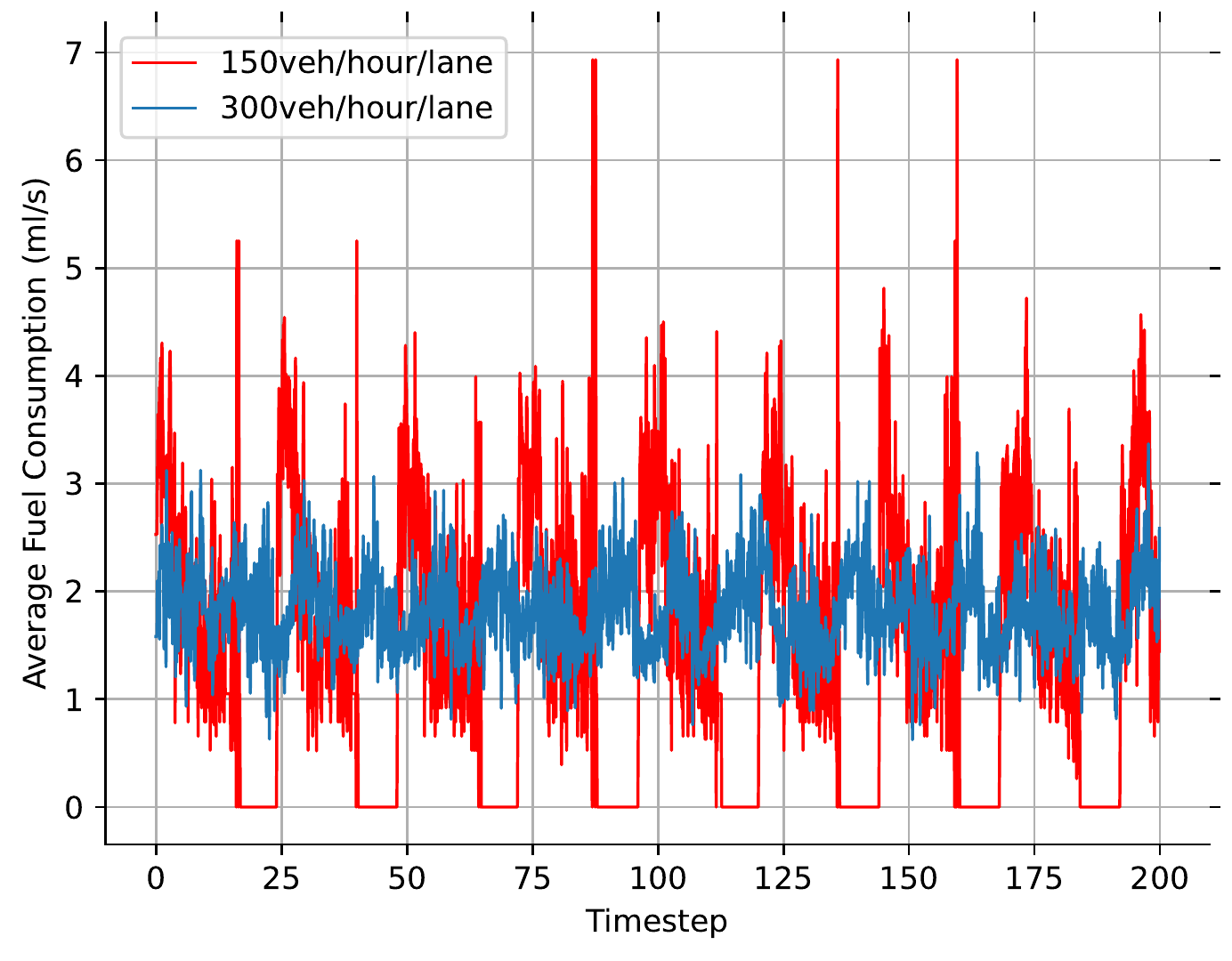}\label{vel_diff_den}}\\
		\subfloat[]{\includegraphics[width=0.46\textwidth]{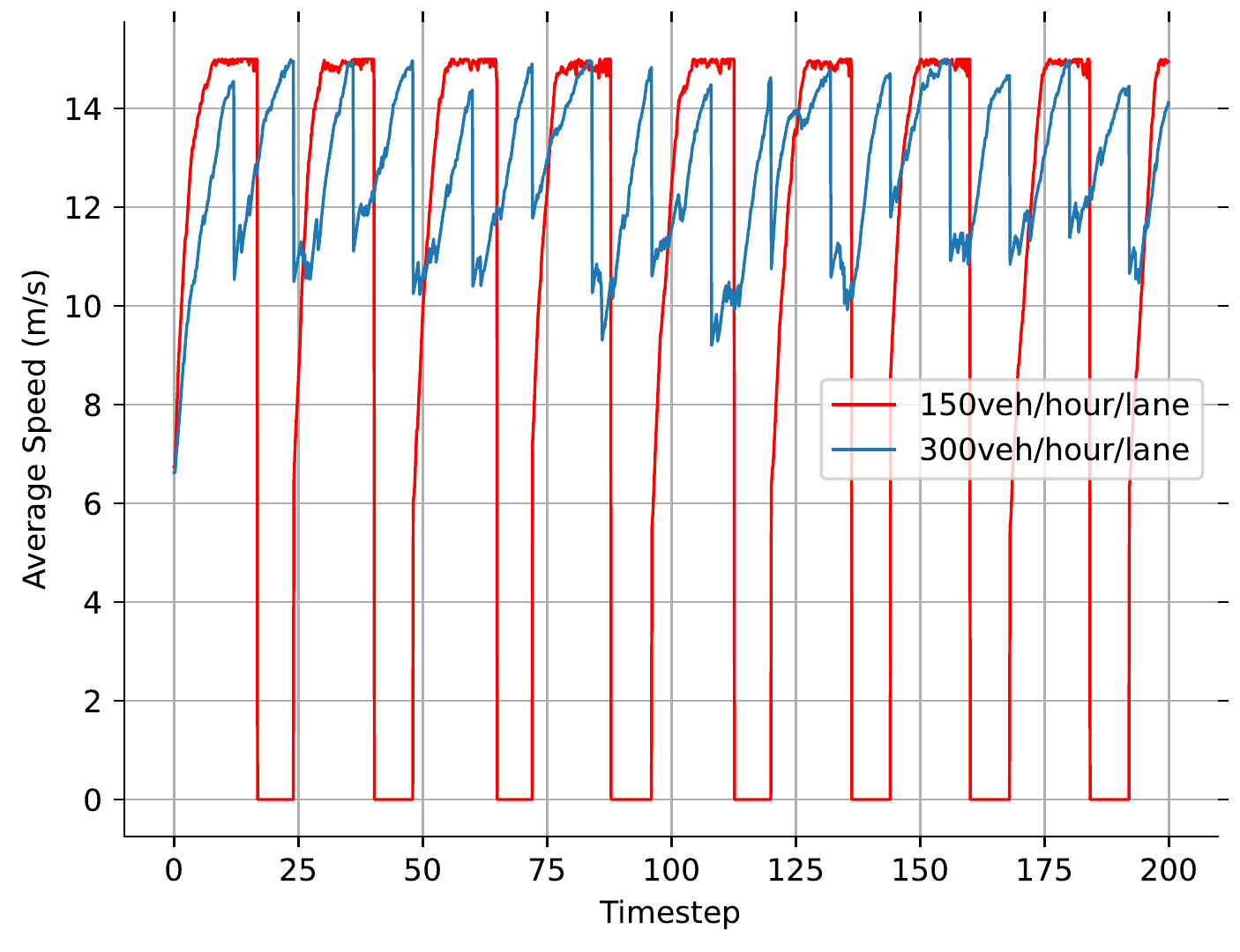}\label{fuel_diff_den}}\\
		\caption{(a) The comparison of the average speed of all vehicles on the road, where CAVs are controlled by the modified QMIX policy, under different densities of traffic flow. (b) The comparison of the average fuel consumption of all vehicles on the road, where CAVs are controlled by the modified QMIX policy, under different densities of traffic flow.}
		\label{eval_diff_den}
	\end{figure}

	We store the model with the maximum reward and the minimum crashes during the training stage as the previous subsection does and compare it with the model under the traffic flow of 150veh/hour/lane using the same three metrics. First, from Fig. \ref{vel_diff_den} and Fig. \ref{fuel_diff_den}, both of the zero plateaus under the traffic density of 150veh/hour/lane disappear since the speed of simulator (SUMO) generating vehicles becomes faster due to the doubled traffic density. However, the speed pattern of more dense traffic is not the same as the scenario with fewer vehicles. Especially in this figure, the rate of speeding becomes lower. The average speed tends to grow from the initial speed to a value slightly lower than the maximum speed of 15m/s and infrequently strikes this maximum value. Subsequently, when there are no vehicles on the road, the average speed value decreases to the random initial speed for the next group of vehicles. Afterward, it rises again, corresponding to the fact that these vehicles coordinate to pass the intersection. Moreover, the vehicles may occasionally decelerate to avoid crashes due to the reward function imposing safety constraints. In Fig. \ref{fuel_diff_den}, the pattern of average fuel consumption becomes noisy in the case of more dense traffic. The spikes with high instantaneous fuel consumption are reduced, resulting from a lower speed-increasing rate and a small gap between the accelerations at consecutive steps.\\
	
	Table. \ref{metrics_compare_tab_300} summarizes the metrics of comparative study between two scenarios. Although the overall average speed decreases when the traffic density doubles, the average fuel consumption is reduced. Unfortunately, few crashes emerge in the dense traffic scenario under the control of our modified QMIX, reflecting the difficulty of generalizing the DRL policy to more realistic traffic scenarios with a large number of vehicles. We leave it to the focus of our future work.
		
	\begin{table}
		\begin{center}
			\caption{Comparison under different densities}
			\label{metrics_compare_tab_300}
			\begin{tabular}{c c c}
				\toprule
				 & \textbf{150veh/hour/lane} & \textbf{300veh/hour/lane}\\
				\midrule
				Average Speed & \textbf{13.52m/s} & 12.61m/s\\
				Average Fuel Consumption & 2.06ml/s & \textbf{1.82ml/s}\\
				Collisions & \textbf{0} & 4 \\
				\bottomrule
			\end{tabular}
		\end{center}
	\end{table}

	\section{Conclusions and Future Works}
	\label{conclusion}
	\noindent
    This paper applies the QMIX algorithm with implementation tricks to coordinate several CAVs at a non-signalized intersection under different densities of mixed-autonomy traffic. First, we conduct a comparative study using the state-of-the-art DRL algorithm, PPO, applied in this area and the original QMIX method under the lower traffic density. The results show that our approach's convergence rate and asymptotic performance beat the others. The model obtained by our algorithm improves the average speed without collisions and coordinates vehicles to leave the intersection with the least fuel consumption. Further, the modified QMIX is trained under the doubled traffic density, where the reward signal eventually converges. Consequently, the stored model can still keep the fuel consumption low, although slightly reducing the efficiency of vehicles and creating marginally more collisions than the lower-traffic counterpart, implying the difficulty of generalizing RL policy to more advanced scenarios.
	
	In the future, we will attempt to assign more explicit credits to each vehicle when they are approaching the intersection by the Shapley value in coalitional game theory, aiming to improve the interpretability of our algorithm. Second, to avoid crashes, we will allow agents to safely explore the environment with higher traffic density, which is consistent with the requirements of real-world applications.

    \section{Acknowledgments}
This work is financially supported by the Institute of Electrical Engineering Chinese Academy of Sciences (E1553301).


	\bibliographystyle{IEEEtran}
	\normalem
	\bibliography{reference.bib}  
	
\begin{IEEEbiography}[{\includegraphics[width=1in,height=1.25in,clip,keepaspectratio]{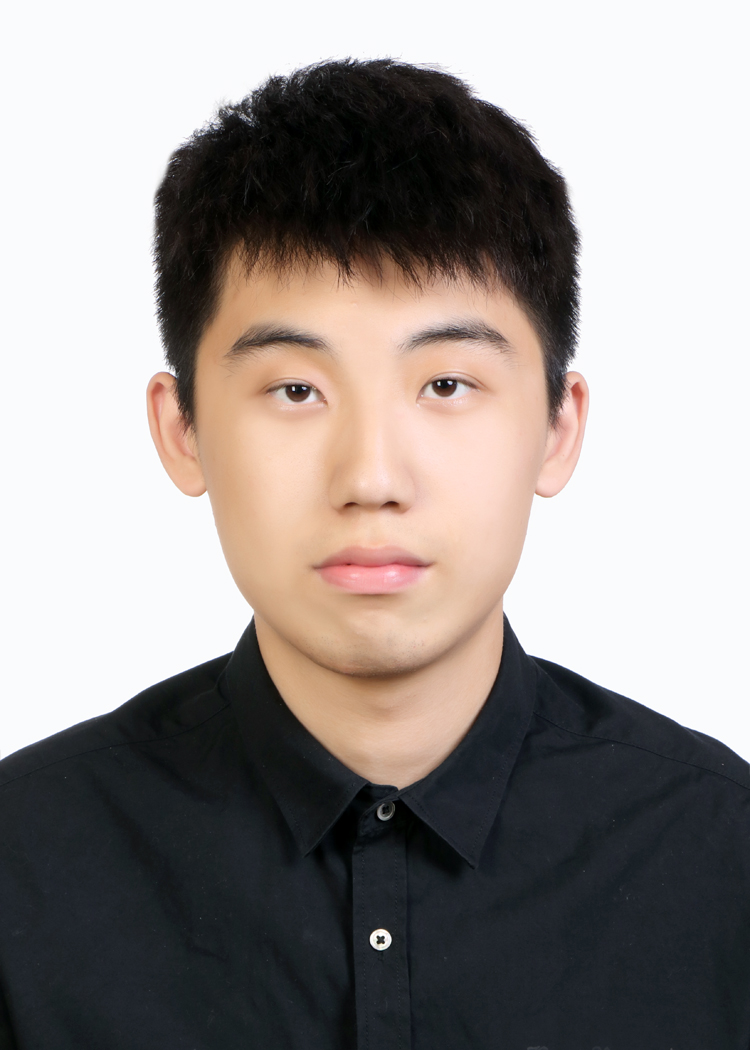}}]{Zihan Guo}
received the B.Eng. degree from Glasgow college, University of Electronic Science and Technology (UESTC), Chengdu, China, and also the B.Eng. degree in Electronics and Electrical Engineering (with Honors of the First Class) from University of Glasgow in 2019. He is pursuing his Ph.D. degree in the Institute of electrical engineering Chinese Academy of Sciences. His research interests include multi-agent deep reinforcement learning, decision making and motion planning for autonomous driving.
\end{IEEEbiography}
\vskip -2\baselineskip plus -1fil

\begin{IEEEbiography}[{\includegraphics[width=1in,height=1.25in,clip,keepaspectratio]{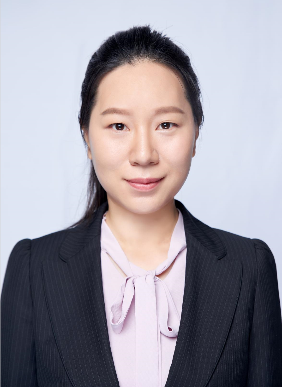}}]{Yan Wu}
received the Ph.D. degree in power electronics and electric drives from the Institute of Electrical Engineering Chinese Academy of Sciences, Beijing, China, in 2019. She is currently an Associate Professor with the Department of Vehicle Energy System and Control Technology Institute of Electrical Engineering Chinese Academy of Sciences. Her research interests include vehicle network information security, X-by-wire systems, vehicle motion control, and intelligent vehicle technology.

\end{IEEEbiography}

\vskip -2\baselineskip plus -1fil

\begin{IEEEbiography}[{\includegraphics[width=1in,height=1.25in,clip,keepaspectratio]{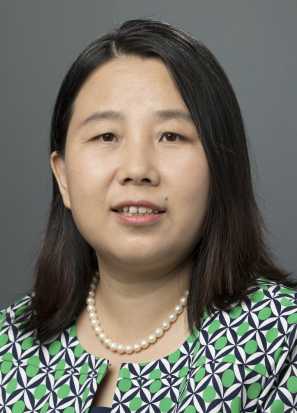}}]{Lifang Wang}
(M’09) received the Ph.D. degree in automobile engineering from Jilin University, Jilin, China, in 1997. She then joined the Institute of Electrical Engineering Chinese Academy of Sciences, Beijing, China. During the Chinese 10th five-year plan (2001–2005), she was a member of the National Specialist Group of the Key Special Electric Vehicle Project of the National 863 Program, and she was the Head of the 863 Special EV Project Office. She is currently the Director of the Department of Vehicle Energy System and Control Technology, Institute of Electrical Engineering Chinese Academy of Sciences, Beijing, China. She is also the Vice Director of the Key Laboratory of Power Electronics and Electric Drives, Chinese Academy of Sciences. Her research interests include electric vehicle control systems, EV battery management systems, wireless charging systems for EVs, electromagnetic compatibility, and smart electricity use. She has directed more than 15 projects in these fields and has published more than 60 papers and 30 patents.

\end{IEEEbiography}

\vskip -2\baselineskip plus -1fil

\begin{IEEEbiography}[{\includegraphics[width=1in,height=1.25in,clip,keepaspectratio]{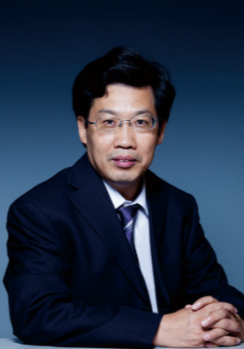}}]{Junzhi Zhang}
(M’09) received the Ph.D. degree in automobile engineering from Jilin University, Jilin, China, in 1997. Since 1998, he has been working in the Department of Automotive Engineering, Tsinghua University, Beijing, China. He is the chairman of the International Electric Vehicle Conference Committee, the expert of the National Energy Conservation and New Energy Vehicle 863 project planning group, and the director of the Electric Vehicle Industry Technology Innovation Alliance. His research interests include Braking energy feedback and dynamic control of the electric vehicle, Energy management and control of hybrid electric vehicles, matching the design of the hybrid electric vehicle, and Design and control of the hybrid transmission.

\end{IEEEbiography}

\vskip -2\baselineskip plus -1fil

\end{document}